# Automatic Open Space Area Extraction and Change Detection from High Resolution Urban Satellite Images

Hiremath P. S.
Dept. of Computer Science
Gulbarga University
Gulbarga, Karnataka State, India

Kodge B.G.
Dept. of Computer Science
S. V. College, Udgir
Dist. Latur, Maharashtra State, India

## ABSTRACT
In this paper, we study efficient and reliable automatic extraction algorithm to find out the open space area from the high resolution urban satellite imagery, and to detect changes from the extracted open space area during the period 2003, 2006 and 2008. This automatic extraction and change detection algorithm uses some filters, segmentation and grouping that are applied on satellite images. The resultant images may be used to calculate the total available open space area and the built up area. It may also be used to compare the difference between present and past open space area using historical urban satellite images of that same projection, which is an important geo spatial data management application.

## 1. INTRODUCTION
Extraction of open space area from raster images is a very important part of GIS features such as GIS updating, geo-referencing and geo spatial data integration. However, extracting open space area from raster image is a time consuming operation when performed manually, especially when the image is complex. The automatic extraction of open space area is critical and essential to the fast and effective processing of large number of raster images in various formats, complexities and conditions.

The method of how open space area are extracted properly from a raster image depends on how open space area appears in the raster image. In this paper, we study automatic extraction of open space area from high resolution urban area satellite image. A high resolution satellite image typically has a resolution of 0.5 to 1.0 m. Under such high resolution, an open space is not same any more in whole image, instead, objects such as lake(s), trees are easily identifiable. This class of images contains very rich information and when fused with vector map can provide a comprehensive view of a geographical area. Google, Yahoo, and Virtual Earth maps are good examples to demonstrate the power of such high resolution images [13]. However, high resolution images pose greater challenges for automatic feature extraction due to the inherent complexities. First, a typical aerial photo captures everything in the area such as buildings, cars, trees, open space area, etc. Second, different objects are not isolated, but are mixed and do interfere with each other, e.g., the shadows of trees on the road, building tops with similar materials. Third, roads, that look like open space area having the same characteristics, cause failure to extract real open space area. In addition, the light and weather conditions have big impact over images. Therefore, it is impossible to predict which and where objects are, and how they look like in a raster image. All these uncertainties and complexities make the extraction process very difficult. Due to its importance, much effort has been devoted to this problem [3, 4]. Unfortunately, there are still no methods that can deal with all these problems effectively and reliably. Some typical high resolution images are shown in Fig.1 and they show huge difference among them in terms of the color spectrum and noise level. There are numerous factors that can distort the edges, including but not limited to blocking objects such as trees and shadows, surrounding objects in similar colors such as roof tops. As a matter of fact, the result of edge detection is as complicated as the image itself. Edges of open space area are either missing or broken, while straight edges correspond to buildings, as shown in Fig.2. Therefore, edge-based extraction schemes do fail to produce reliable results under such circumstances.

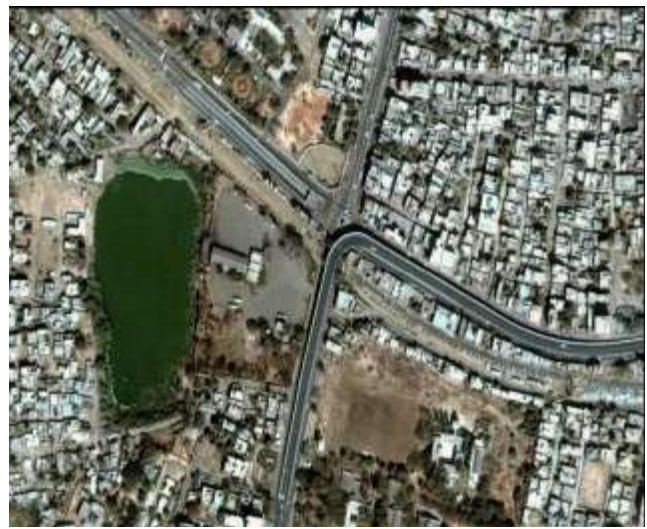

(a)

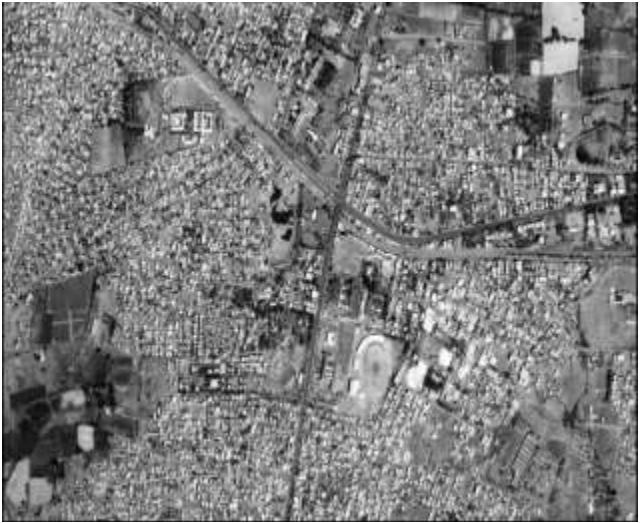

**(b)**

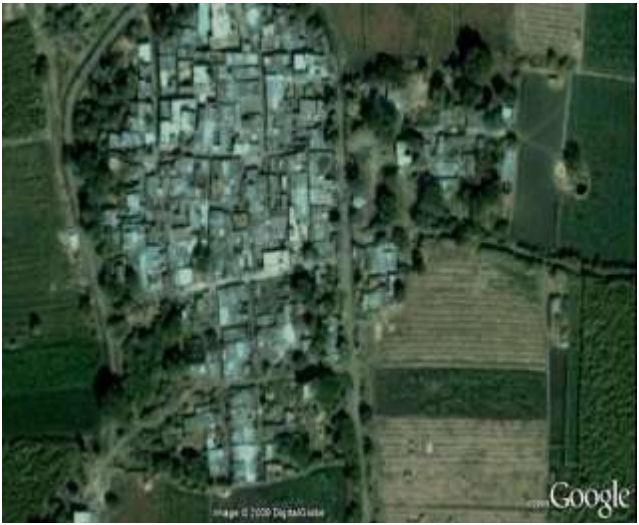

**(c)**

Figure 1. Examples of high resolution images (a) An aerial photo of an area in the Latur city, (b) High resolution urban satellite image of Latur city (dated 23 Feb. 2003), (c) Google map satellite image.

In this paper, we develop an integrated scheme for automatic extraction that exploits the inherent nature of open space area. Instead of relying on the edges to detect open space, it tries to find
the pixels that belong to the same open area region based on how they are related visually and geometrically. Studies have shown that the visual characteristic of a open space is heavily influenced by its physical characteristics such as material, surface condition. It is impossible to define a common pattern just based on color or spectrum to uniquely identify an open space area. In our scheme, we consider an open space as a group of "similar" pixels. The similarity is defined in the overall shape of the region they belong to, the spectrum they share, and the geometric property of the region. Different from edge-based extraction schemes, the new scheme first examines the visual and geometric properties of pixels using a new method. The pixels are identified to represent each region. All the regions are verified based on the general visual and geometric constraints associated with a open space area. Therefore, the roof top of a building or a long strip of trees is not misidentified as a open space area segment. There is also no need to assume or guess the color spectrum of open space area, which varies greatly from image to image, as evident from the example images.

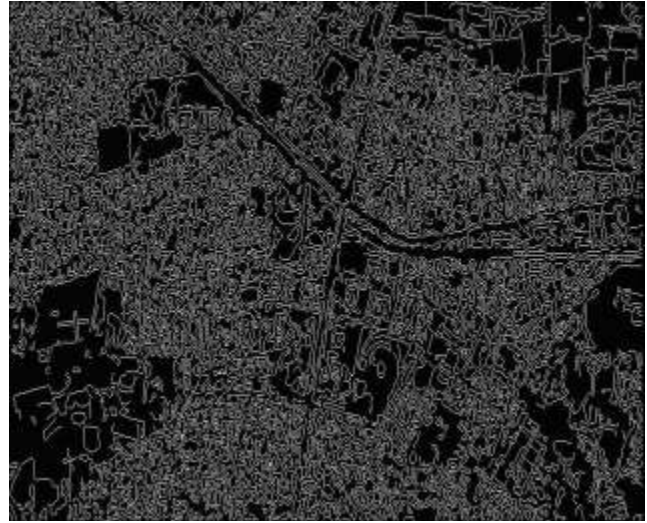

Figure 2. Edge extraction (using canny method) of Fig.1(b).

As illustrated by examples in Fig.1, an open space area is not always a contiguous region of regular linear shape, but a series of segments with non regular shapes. This is because each segment may include pixels of surrounding objects in similar colors or miss some of its pixels due to interference of surrounding objects. A reliable extraction scheme must be able to deal with such issues. In the following sections, we discuss how to capture the essence of "similarity", translate them and finally turn them into display.

## 2. IDENTIFICATION OF WATER BODIES FROM SATELLITE IMAGES

While extracting open space area from high resolution urban satellite images, first, we should identify the water bodies like lakes or reservoirs or tanks or rivers etc. For this purpose, a large number of times the earth observation satellites have orbited, and are orbiting, our planet to provide frequent imagery of earth's surface. From these satellites, many images can potentially provide useful information for assessing erosion, although less has actually been used for this purpose. This section provides a brief overview of the space borne sensors applied in water-body extraction studies. Optical satellite systems are, most frequently, applied in water body extraction research. The parts of the electromagnetic spectrum covered by these sensors include the visible and near infrared (VNIR) ranging from 0.4 to 1.3 μm, the shortwave infrared (SWIR) between 1.3 and 3.0 μm, the thermal infrared (TIR) from 3.0 to 15.0 μm and the long-wavelength infrared (LWIR) from (7-14 μm). Table 1 summarizes sensor characteristics of the systems used.[12].

**Table 1: The optical satellite sensors used for water body extraction**

| Satellites | Sensors | Operation Time | Spatial Resolution |
|---|---|---|---|
| Landsat-1,2,3 | MSS | 1972-1983 | 80m |
| NOAA/TIROS | AVHRR | 1978-Present | 1001m |
| Nimbus-7 | CZCS | 1978-1986 | 825m |
| Landsat-4,5 | TM | 1982-1999 | 120m |
| SPOT-1,2,3 | HRV | 1986-Present | 20m |
| IRS- 1A,1B | LISS-1 LISS-2 | 1988-1999 | 72.5m 36.25m |
| IRS-1C,1D | PAN | 1995-Present | 5.8m |
| SPOT-4 | HRVIR | 1998-Present | 10m |
| IKONOS | Panchromatic Multispectral | 1999-Present | 1m 2m |
| Landsat-7 | ETM | 1999-Present | 15m 30m |
| TERRA | ASTER MODIS | 1999-Present 1999-Present | 15m 250m |
| QuickBird | Panchromatic Multispectral | 2001-Present | 0.61m |
| SPOT-5 | Panchromatic Multispectral | 2002-Present | 5m |
| WorldView-1 | Panchromatic | 2007-Present | 0.55m |
| GEOEYE-1 | Pan-sharpened Panchromatic Multispectral | 2008-Present | 0.41m |

## 3. AUTOMATIC EXTRACTION SCHEME

The goal of the proposed system is to develop a complete and practical solution to the problem of automatic extraction of open space area in high resolution aerial/satellite images.

In the first stage, we extract open space areas that are relatively easier to identify such as major grounds and, in the second stage, we deal with open space that are harder to identify. The reason for such design is a balance between reliability and efficiency. Some open space areas are easier to identify, because they are more identifiable and contain relatively less noise. Since some open space area in the same image share some common visual characteristics, the information from the already extracted area and other objects, such as spectrum, can be used to simplify the process of identifying open space area that are less visible or heavily impacted by surrounding objects or by different colors. Otherwise, such areas are not easily distinguishable from patterns formed by other objects. For example, a set of collinear blocks may correspond to a open space or a group of buildings (houses) from the same block. The second stage also serves an important purpose to fill the big gaps for open space extracted in stage one. Under some severe noise, part of the area may be disqualified as valid open space region and hence missed in stage one, leaving some major gaps in open space area edges. With the additional spectrum information, these missed areas can be easily identified to complete the open space area extraction. Therefore, the two stage process eliminates the need to assume or guess the color spectrum of open space and allow extraction much more completely.

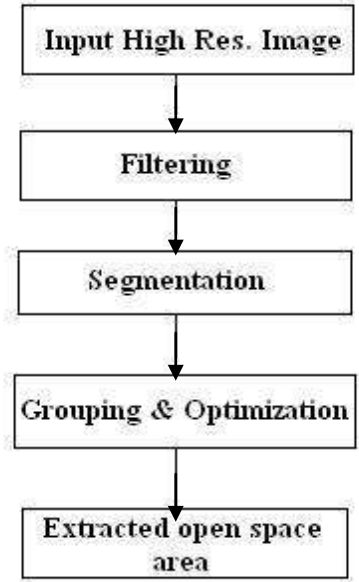

Figure 3. Automatic extraction scheme.

Each major stage consists of the following three major steps: filtering, segmentation and, grouping and optimization, as shown in Fig.3. The details of each step will be discussed in the following section.

## 4. ALGORITHM

We assume that the images satisfy the following two general constraints, which are derived on the basis of minimum conditions for an open space area to be identifiable and, therefore, are easily met by most images: (i) Visual constraint: majority of the pixels from the same open space area have similar spectrum that is distinguishable from most of the surrounding areas; and, (ii)Geometric constraint: a open space is a region that has relatively no standard shape, compared with other objects in the image.

These two constraints are different in implications. The visual constraint does not require a open space region to have single color or constant intensity. It only requires blank area to look visually different from surrounding objects in most parts. The geometric constraint does not require a smooth edge, only the overall shape to be a long narrow strip. So these conditions are much weaker and a lot more practical. As we can see, these assumptions can accommodate all the difficult issues very well, including blurring, broken or missing edge of open area boundaries, heavy shadows, and interfering surrounding objects.

## 4.1 Filtering

The step of filtering is to identify the key pixels that will help determine if the region they belong to is likely an open space area segment. Based on the assumption of visual constraint, it is possible to establish an image segmentation using methods of edge detection. We note that such separation of regions is required to be precise and normally suffers from quite a lot of noise. In the best, the boundaries between regions are a set of line segments for most images, as is the case shown in Fig.2. It certainly does not tell which region corresponds to which area and which is not based on the extracted edges. As a matter of fact, most of the regions are not completely separated by edges and are still interconnected on 4-connect path or 8-connect path. In order to fully identify and separate open space regions from the rest of image, we propose to invert them and again extract the edges using Sobel edge detector to highlight sharp changes in intensity in the active image or selection. Two 3x3 convolution kernels (shown below) are used to generate vertical and horizontal derivatives. The final image is produced by combining the two derivatives using the square root of the sum of the squares.

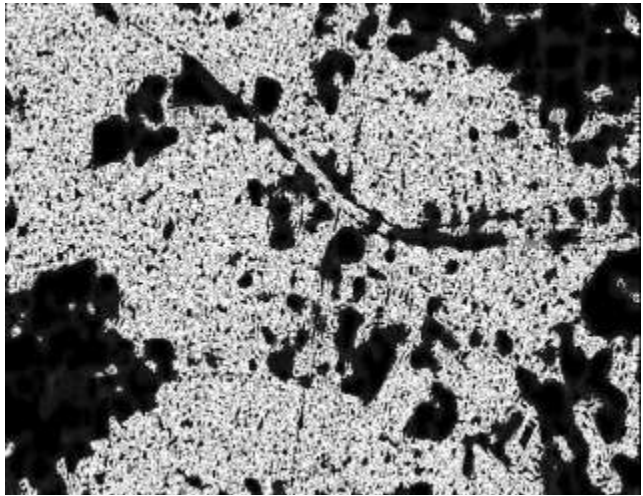

Figure 4. Resultant image after edge detection and outlier removal in image of Fig.2

```
 1  2  1     1  0 -1
 0  0  0     2  0 -2
-1 -2 -1     1  0 -1
```

The next step is a removal of outliers through a process, namely: we replace a pixel by the median of the pixels in the surrounding if it deviates from the median by more than a certain value (the threshold). We used the following values for outlier removal.

R = 10.0 pixels, T = 2, OL = Bright.

The R is radius that determines the area used for calculating the median (uncalibrated, i.e., in pixels). The Fig.4 depicts how radius translates into an area. The T (Threshold) determinates by how much the pixel must deviate from the median to get replaced, in raw (uncalibrated) units. The outlier OL determines whether pixels brighter or darker than the surrounding (the median) is replaced.

## 4.2 Segmentation

The step of segmentation is to verify which region is a possible road region based on the central pixels. Central pixels contain not only the centerline information of a region, but also the information of its overall geometric shape. For example, a perfect square will only have one central pixel at its center. A long narrow strip region will have large number of central pixels. Only regions with ratios above certain threshold are considered to be candidate regions. In order to filter out interference as much as possible for reliable extraction during the first major stage, a minimum region width can be imposed. This will effectively remove most of the random objects from the image. However, such width constraint will be removed during the second major step as improper regions can also be filtered out based on the color spectrum information obtained from stage one. Therefore, small regions with closer spectrum are examined for possible small open space area in the second stage.

In addition to the geometry constraint, some general visual information can also be applied to filter out obvious non-open space regions. For example, if the image is a color image, most of the tree and grass areas are greenish. Also tree areas usually contain much richer textures than normal smooth road surfaces. The intensity transformation and spatial and frequency filtering can be used to filter out such areas. The minimum assumption, namely, small regions with closer spectrum, of the proposed scheme does not exclude the use of additional visual information to further improve the quality of extraction if they are available.

## 4.3 Grouping and Optimization

As the result of segmentation, the open space area segments are typically incomplete and disconnected due to heavy noise.

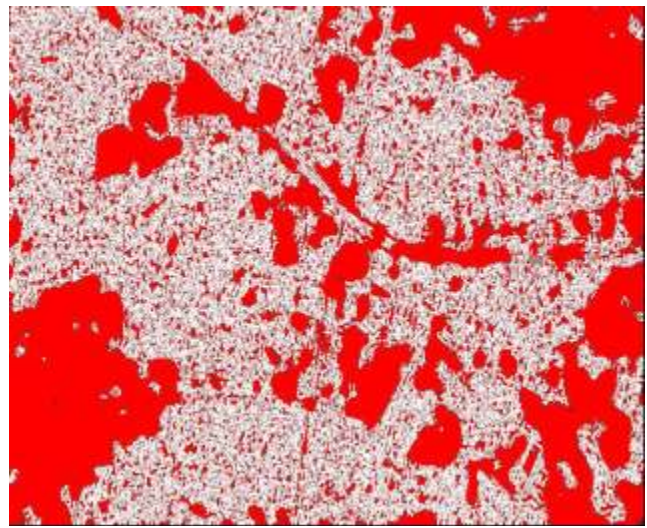

Figure 5. Result after applying upper and lower threshold values.

The purpose of this step is to group corresponding open space area segments together in order to find the optimal results for required area extraction. If enough information is available to determine the open space area spectrum, then optimization is better applied after all the segments are identified. The Fig.5 is the result of thresholding, based on the automatically or interactively set lower=0 and upper=48 threshold values, which segments the image into features of interest and background. The features of interest are displayed in white. The background is displayed in red color, which is the total open space area in the given projected satellite image.

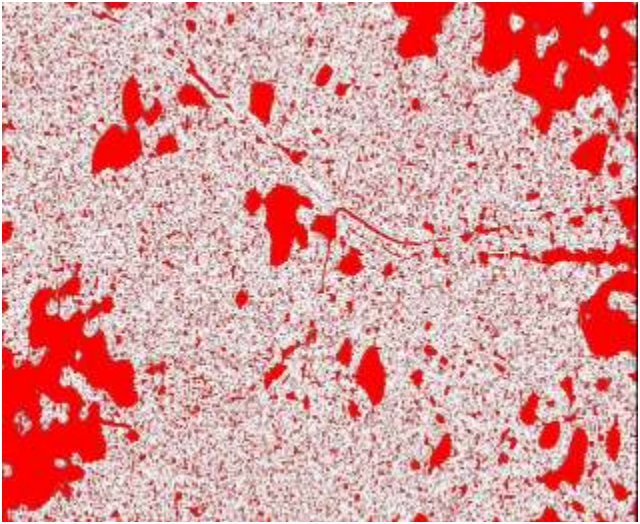

(a)

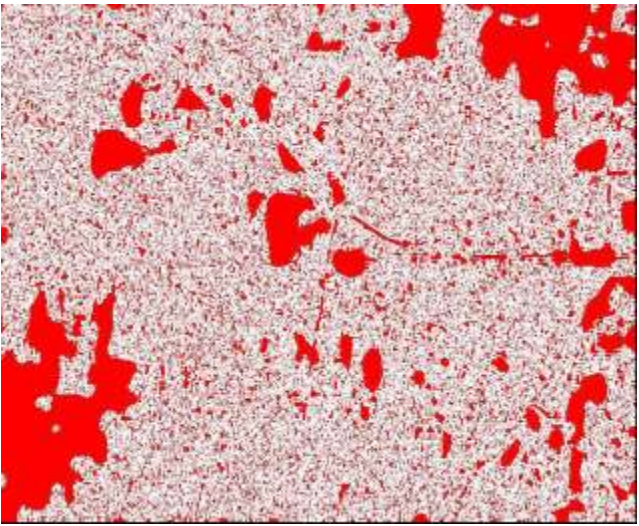

(b)

Figure 6. Extracted open space area (Red color) using available historical images (a) Available open space area on Feb 18 2006 (b) Available open space area on Jan 11 2008.

The open space area extracted from high resolution urban satellite imagery is shown in the Fig.7 with labeled regions. The area and centroid of each labeled region of the total available open space in Figs.7-9 are shown in the Tables 2-4, respectively.

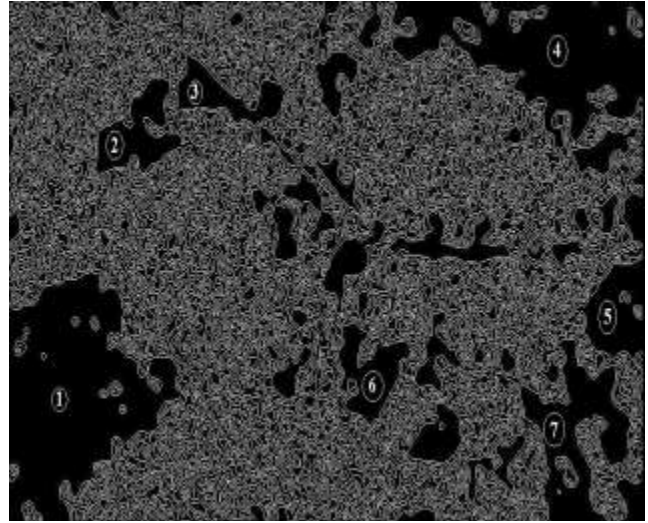

Figure 7 Labeled regions of extracted open space area (Feb. 2003)

**Table 2. Area and centroids of labeled regions of Fig.7.**

| Labels | Area | Centroids | |
|---|---|---|---|
| | | x1 | y1 |
| 1 | 33757 | 414.899 | 265.154 |
| 2 | 4079 | 446.027 | 281.925 |
| 3 | 3559 | 437.167 | 261.494 |
| 4 | 28865 | 488.196 | 248.630 |
| 5 | 5890 | 455.214 | 276.307 |
| 6 | 4368 | 445.361 | 271.004 |
| 7 | 11221 | 449.157 | 282.151 |

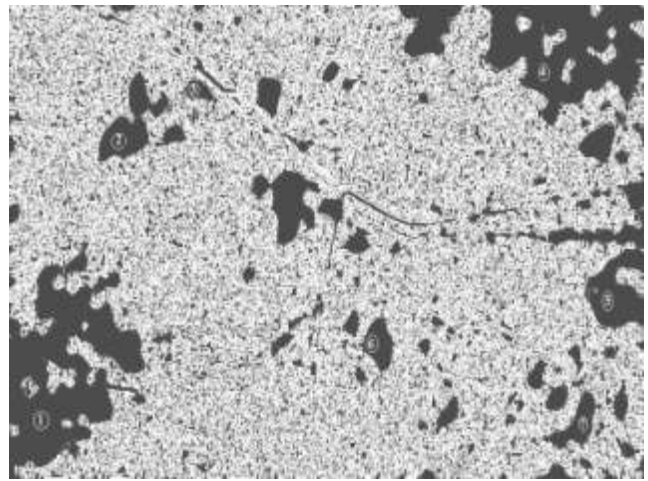

Figure 8. Labeled regions of extracted open space area (Feb. 2006).

**Table 3. Area and centroids of labeled regions of Fig.8.**

| Labels | Area | Centroids | |
|---|---|---|---|
| | | x1 | y1 |
| 1 | 18191 | 67.115 | 432.29 |
| 2 | 2871 | 172.95 | 138.836 |
| 3 | 647 | 270.296 | 99.721 |
| 4 | 18596 | 792.175 | 49.448 |
| 5 | 4460 | 860.27 | 332.954 |
| 6 | 1286 | 524.452 | 389.514 |
| 7 | 1708 | 805.73 | 484.958 |

From the Tables 2-4, the comparative study of open space areas, extracted from existing historical images of Latur city in the years 2003, 2006 and 2008, is demonstrated graphically in the Fig.10.

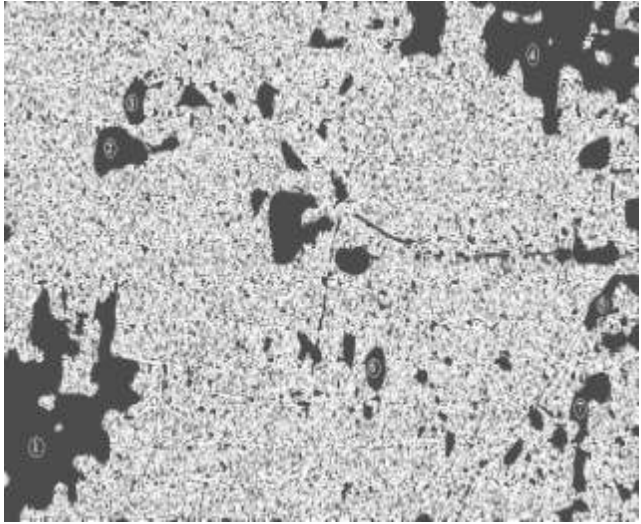

Fig.9 Labeled regions of extracted open space area (Jan 2008).

**Table 4. Area and centroids of labeled regions of Fig.9.**

| Labels | Area | Centroids | |
|---|---|---|---|
| | | x1 | y1 |
| 1 | 15767 | 67.211 | 440.465 |
| 2 | 2461 | 166.916 | 155.971 |
| 3 | 407 | 266.52 | 101.181 |
| 4 | 10561 | 777.8 | 49.802 |
| 5 | 1289 | 855.495 | 312.981 |
| 6 | 960 | 525.872 | 386.453 |
| 7 | 1764 | 829.141 | 419.818 |

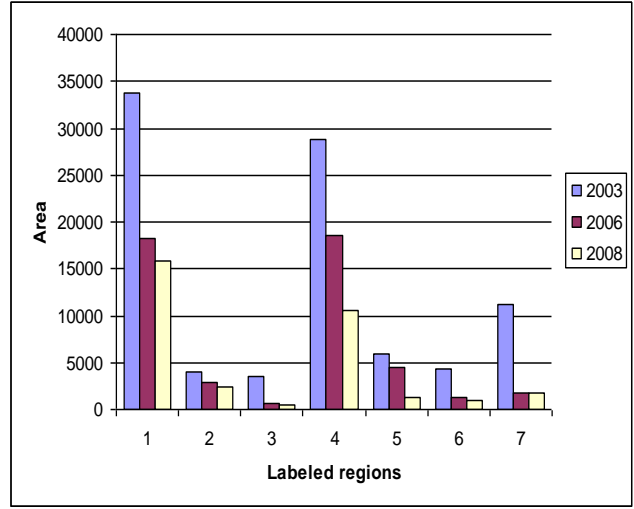

Figure 10. Comparative results of extracted open space areas of imageries of Latur city in the years 2003, 2006 and 2008.

## 5. Change Detection

Change detection is used to highlight or identify significant differences in imagery acquired at different times and thus plays an important role in the lifecycle of GIS. Based on the results of above experiment, the Figs.11, 12 and 13 show the changes detected during the periods 2003 to 2008, 2003 to 2006 and 2006 to 2008 from the 2003, 2006 and 2008 year's satellite imageries.

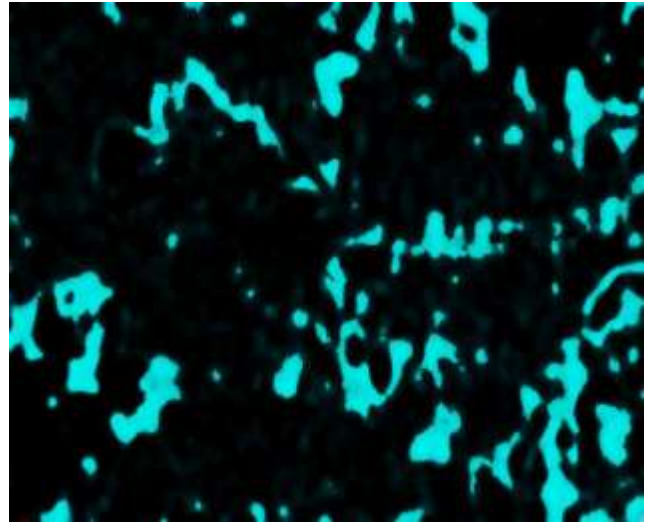

Figure 11. Changes detected during 2003 to 2008

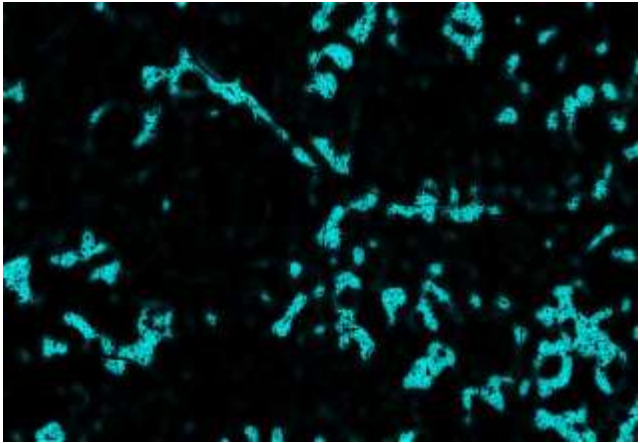

Figure 12. Changes detected during 2003 to 2006.

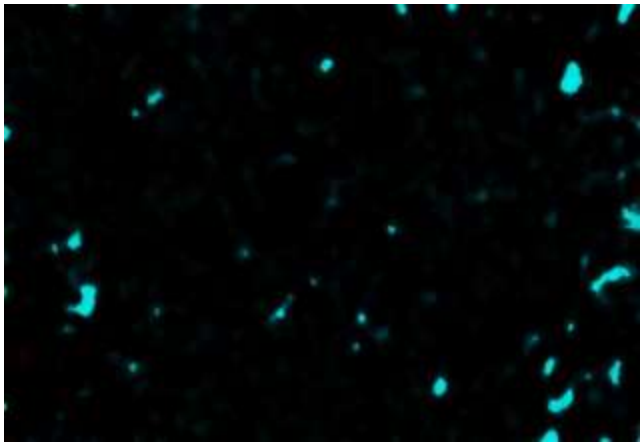

Figure 13. Changes detected during 2006 to 2008.

## 6. CONCLUSIONS

In this paper, we have proposed a new automatic system for extracting open space area and change detection during a certain time period. The main contribution of the proposed system is to address the major issues that have caused all existing extraction approaches to fail, such as blurring boundaries, interfering objects, inconsistent area profiles, heavy shadows, etc. To address all these difficult issues, we develop a new method, namely automatic extraction of open space area from high resolution satellite imagery, to capture the essence of both visual and geometric characteristics of open space area. The extraction process including filtering, segmentation, and grouping and optimization, eliminates the need to assume or guess the color spectrum of different open space areas. The proposed approach is efficient, reliable, and assumes no prior knowledge about the required open space area, conditions and surrounding objects. It is able to process complicated aerial/satellite images from a variety of sources including aerial photos from Google and Yahoo satellite images. The quick application of the proposed study is envisaged to help landing the helicopters in open space area safely in case of emergencies, urban development planning, land use/and land cover assessment.

## 7. ACKNOWLEDGEMENT

Authors are grateful to referees for their valuable comments.